\pdfoutput=1

\documentclass[11pt]{article}

\usepackage{stylesheets/emnlp2021/emnlp2021}

\usepackage{times}
\usepackage{latexsym}

\usepackage{xcolor}

\usepackage[T1]{fontenc}

\usepackage[utf8]{inputenc}

\usepackage{microtype}

\usepackage[textsize=scriptsize]{todonotes}
\usepackage{csquotes}
\MakeOuterQuote{"} 
\usepackage{tikz}
\usetikzlibrary{positioning, calc, shapes, arrows}
\usepackage{amsmath}
\usepackage{amssymb}
\usepackage{mathtools}

\usepackage{algorithm}  
\usepackage{algorithmic}
\usepackage{booktabs}
\usepackage{multirow}
\usepackage{placeins} 
\usepackage{CJKutf8} 
\usepackage{bbding}

\usepackage{stfloats}

\renewcommand{\sec}{\S}
\newcommand{\tab}{Tab.~} 
\newcommand{\fig}{Fig.~}

\usepackage{rotating}

\title{Towards Label-Agnostic Emotion Embeddings}

\author{
Sven Buechel \hspace*{.5cm}
Luise Modersohn \hspace*{.5cm}
Udo Hahn\\
 Jena University Language \& Information Engineering (JULIE) Lab\\
 Friedrich-Schiller-Universit\"at Jena, F\"urstengraben 27, 07743 Jena, Germany \vspace*{5pt} \\
 {\tt firstname.lastname@uni-jena.de}\\
{\tt\href{https://julielab.de}{https://julielab.de}}\\
}

\begin{document}
\maketitle
\begin{abstract}
Research in emotion analysis is scattered across different label formats  (e.g., polarity types, basic emotion categories, and affective dimensions), linguistic levels (word vs.\ sentence vs.\ discourse), and, of course, (few well-resourced but much more under-resourced) natural languages and text genres (e.g., product reviews, tweets, news).\ The resulting heterogeneity makes data and software developed under these conflicting constraints hard to compare and challenging to integrate. To resolve this unsatisfactory state of affairs we here propose a training scheme that learns a shared latent representation of emotion independent from different label formats, natural languages, and even disparate model architectures. Experiments on a wide range of datasets indicate that this approach yields the desired interoperability without penalizing prediction quality. Code and data are archived under DOI \href{https://doi.org/10.5281/zenodo.5466068}{\tt \small 10.5281/zenodo.5466068}.
\end{abstract}

\section{Introduction} \label{sec:intro}

\begin{table*}[t]
	\footnotesize
	\centering
	\setlength{\tabcolsep}{3pt}
    \newcommand{\lightcircle}{\raisebox{1mm}{$\circ$}}
\newcommand{\darkcircle}{\raisebox{1mm}{$\bullet$}}
\newcommand{\lighttriangle}{{\tiny\textsuperscript{$\triangle$}}}
\newcommand{\lightsquare}{{\tiny\textsuperscript{$\square$}}}
\newcommand{\darksquare}{{\tiny\textsuperscript{$\blacksquare$}}}

	\begin{tabular}{p{7.5cm}|ccc|ccccc}
	\toprule
	\textbf{Sample} &   \textbf{Val} & \textbf{Aro} & \textbf{Dom} & \textbf{Joy} & \textbf{Ang} & \textbf{Sad} & \textbf{Fea} & \textbf{Dis} \\
    \midrule
		rollercoaster  	& 8.0\lightcircle & 8.1\lightcircle & 5.1\lightcircle & 3.4\lightsquare & 1.4\lightsquare	 & 1.1\lightsquare	& 2.8\lightsquare	& 1.1\lightsquare\\
		urine &	 3.3\lightcircle& 4.2\lightcircle & 5.2\lightcircle & 1.9\lightsquare & 1.4 \lightsquare & 1.2\lightsquare & 1.4\lightsquare &	2.6\lightsquare	 \\
		szczęśliwy \textsuperscript{(a)}  & 2.8\darkcircle & 4.0\lightcircle & & &&&& \\%
		\midrule
		College tution continues climbing &&& &0\darksquare&54\darksquare & 40\darksquare & 3\darksquare & 31\darksquare \\ 
		A gentle, compassionate drama about grief and healing  & $pos$\lighttriangle &&&&&&& \\
		\begin{CJK*}{UTF8}{bsmi}喇叭這一代還是差勁透了。\end{CJK*} \textsuperscript{(b)}  & 2.8\lightcircle & 6.1\lightcircle & & & & & &  \\ 
		\bottomrule
	\end{tabular}\vspace{.1cm}
	
    Value Ranges: \hspace{1cm} \lightcircle $[1,9]$ \hspace{1cm} \darkcircle $[-3,3]$ \hspace{1cm} \lighttriangle $\{pos, neg\}$ \hspace{1cm} \lightsquare $[1, 5]$ \hspace{1cm} \darksquare $[0,100]$

	\caption{Sample entries from various sources described along eight emotional variables:\newline  [VAD]---\textbf{Val}ence ($\approx$ \textbf{Pol}arity), \textbf{Aro}usal, \textbf{Dom}inance, and [BE5]---\textbf{Joy}, \textbf{Ang}er, \textbf{Sad}ness, \textbf{Fea}r, and \textbf{Dis}gust. \newline
	Samples differ in languages addressed (English, Polish, Mandarin), linguistic domain (word vs.\ text, register) and label format (covered variables and their value ranges). \newline
	Translations: \textsuperscript{(a)} \enquote{happy} (from Polish); \mbox{\textsuperscript{(b)} \enquote{This product generation still has terrible speakers.} (from Mandarin)}
		\label{tab:examples}}
    \vspace*{-3pt}
\end{table*}

Emotion analysis in the field of NLP\footnote{We use \enquote{emotion} as an umbrella term for phenomena such as polarity, sentiment, feelings, or affective states.}  has experienced a remarkable evolution of representation schemes. Starting from the early focus on \textit{polarity}, i.e., the main distinction between positive and negative feelings emerging from natural language utterances \citep{Hatzivassiloglou97acl,Turney03}, the number and variety of label formats, i.e., groups of emotional target variables and their associated value ranges, has been growing rapidly \cite{Bostan18coling,DeBruyne20lrec}.  This development is a double-edged sword though.

On the one hand, the wide variety of available label formats allows NLP models to become more informative and richer in expressive power. This gain is because many of the newer representation schemes follow well-researched branches of psychological theory, such as basic emotion categories or affective dimensions \citep{Ekman92,Russell77}, which offer information complementary to each other \cite{Stevenson07}. Others argue that different emotional nuances turn out to be particularly useful for specific targeted downstream applications \citep{Bollen11,Desmet13}.

On the other hand, this proliferation of label formats has led to a severe loss in cross-data comparability. As \tab\ref{tab:examples} illustrates, the total volume of available gold data is spread not only over distinct languages but also a huge number of emotion annotation schemes. Consequently, comparing or even merging data from different rating studies is often impossible. This, in turn, contributes to the development of an unnecessarily large number of prediction models, each with limited coverage of the full range of human emotion.

To escape from these dilemmata, we propose a method that mediates between such different representation schemes. In contrast to previous work which unified \textit{some} sources of heterogeneity (see \sec\ref{sec:related}), to the best of our knowledge, our approach is the first to learn a representation space for emotions that \textit{generalizes} over individual languages, emotion label formats, and distinct model architectures for emotion analysis. 

\begin{figure}[t]
    \centering
    \includegraphics[width=.5\textwidth]{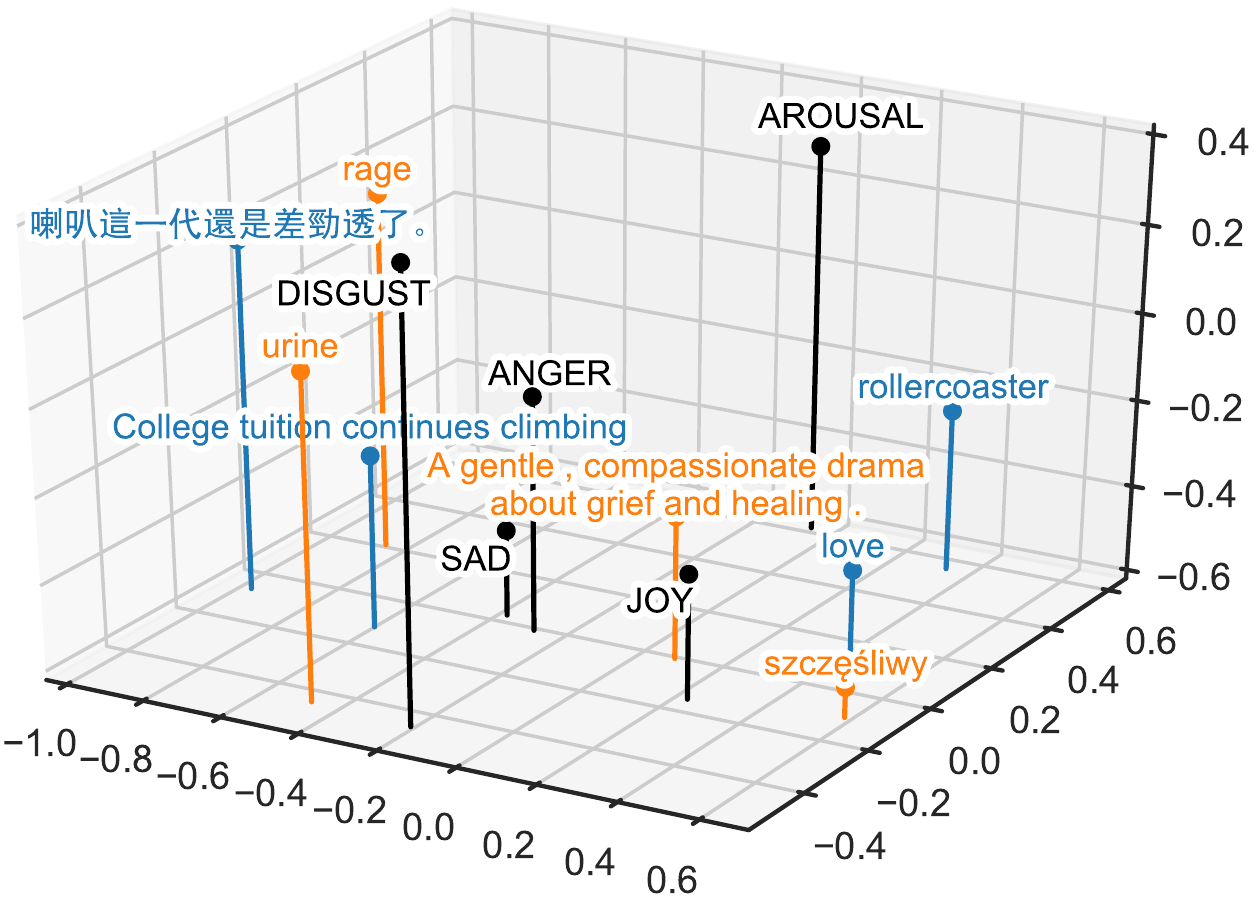}
    \caption{Emotional loading of heterogenous samples in common representation space with selected emotion variables (in capitals); first three principal components. Color only used as visual aid. Translations for non-English items are given in \tab\ref{tab:examples}.}
    \label{fig:emotion-space}
\end{figure}

Technically speaking, our approach consists of a set of pre-trained prediction heads that can be easily attached to existing state-of-the-art neural models. Doing so, a model learns to \textit{embed} language items of a particular domain in a shared representation space that resembles an "interlingua for emotion".  These "emotion embeddings" capture a rich array of affective nuances and allow for a direct comparison of emotional load between heterogeneous samples (see \fig\ref{fig:emotion-space}). They may thus form a solid basis for a broad range of linguistic, psychological, and cultural follow-up studies. 

In terms of practical benefits, our method allows models to predict label formats unseen during training and lowers space requirements by reducing a large number of format-specific models to a small number of format-agnostic ones. Although not in the center of interest of this study, our approach also often leads to small improvements in prediction quality, as experiments on 13 datasets for 6 natural languages reveal.

\section{Related Work}
\label{sec:related}

\paragraph{Representing Emotion.}

At the heart of computational emotion representation lies a set of \textit{emotion variables} (\enquote{classes}, \enquote{constructs}) used to capture different facets of affective meaning. Researchers may choose from a multitude of approaches designed in the long and controversial history of the psychology of emotion \citep{Scherer00,hofmann_appraisal_2020}.  A popular choice are  so-called \textit{basic emotions} \cite{Alm05,Aman07,Strapparava07}, such as the six categories identified by \citet{Ekman92}: \textit{Joy}, \textit{Anger}, \textit{Sadness}, \textit{Fear}, \textit{Disgust}, and \textit{Surprise} (\textbf{BE6}, for short). A subset of these excluding \textit{Surprise} (\textbf{BE5}) is often used for emotional word datasets in psychology (\enquote{affective norms}) which are available for a wide range of languages.

\textit{Affective dimensions} constitute a popular alternative to basic emotions \citep{Yu16naacl,Sedoc17eacl,Buechel17eacl,Li17,Mohammad18acl}. The most important ones are \textit{Valence} (negative vs.\ positive, thus corresponding to the notion of \textit{polarity}; \citealp{Turney03}) and \textit{Arousal} (calm vs.\ excited) (\textbf{VA}). These two dimensions are  sometimes extended by \textit{Dominance} (feeling powerless vs.\ empowered; \textbf{VAD}).    

Other theories influential for NLP include Plutchik's (\citeyear{plutchik_nature_2001})  \textit{Wheel of Emotion} \citep{Mohammad13,Abdul17acl,Tafreshi18,Bostan20lrec} and appraisal dimensions \citep{Balahur12,Troiano19acl,hofmann_appraisal_2020}. Yet frequently, studies do not follow any of these established approaches but rather design a customized set of variables in an ad-hoc fashion, often driven by the availability of user-labeled data in social media, or the specifics of an application or domain which requires attention to particular emotional nuances \citep{Bollen11,Desmet13,Staiano14,Qadir14emnlp,Li16,Demszky20acl}. 

This proliferating diversity of emotion label formats is the reason for the lack of comparability outlined in \sec \ref{sec:intro}. Our work aims to unify these heterogeneous labels by learning to translate them into a shared distributional representation (see \fig \ref{fig:emotion-space}).

\paragraph{Analyzing Emotion.}

There are several subtasks in emotion analysis that require distinct model types. 
Word-level prediction (or \enquote{emotion lexicon induction}) is concerned with the emotion associated with an individual word out of context. Early work exploited primarily surface patterns of word usage \citep{Hatzivassiloglou97acl,Turney03} whereas more recent activities rely on more sophisticated statistical signals encoded in word embeddings \citep{Amir15,Rothe16,Li17}. Combinations of high-quality embeddings with feed-forward nets have proven to be very successful, rivaling human annotation capabilities \citep{Buechel18naacl}.

In contrast, modeling emotion of sentences or short texts (jointly referred to as \enquote{text})  was traditionally based largely on lexical resources \cite{Taboada11cl}. Later, those were combined with conventional machine learning techniques \cite{Mohammed13semeval} before being widely replaced by neural end-to-end approaches \citep{Socher13,Kim14emnlp,Abdul17acl}. Current state-of-the-art results are achieved by transfer learning with transformer models \citep{Devlin19naacl,Zhong19,Delbrouck20}.

Our work complements these lines of research by providing a method 
that allows existing models to embed the emotional loading of some unit of language in a common emotion embedding space. This broadens the range of emotional nuances said models can capture. Importantly, our method learns a representation not for a specific unit of language itself but the emotion attached to it. This differs from previous work aiming to increase the affective load of, e.g., word embeddings (see below).

\paragraph{Emotion Embeddings.}
Several existing studies have used the term "emotion embeddings" (or similar phrasing) to characterize their work, yet either use the term in a different way or tackle a different problem compared to our study.

In more detail, \citet{Wang20tit} present a method for increasing the emotional content of word embeddings based on re-ordering vectors according to the similarity in their emotion values, referring to the result as "emotional embeddings". Similarly, \citet{Xu18wassa} learn word embeddings that are particularly rich in affective information by sharing an embedding layer between models for different emotion-related tasks. They refer to these embeddings as "generalized emotion representation". 
Different from our work, these two studies primarily learn to represent \textit{words} (with a focus on their affective meaning though), not emotions themselves. They are thus in line with previous research aiming to increase the affective load of word embeddings \citep{Faruqui15naacl,Yu17emnlp,Khosla18}.

\citet{Shantala18} improve a dialogue system by augmenting their training data with emotion predictions from a separate system. Predicted emotion labels are fed into the dialogue model using a representation ("emotion embeddings") learned in a supervised fashion with the remainder of the model parameters. These embeddings are specific to their architecture and training dataset, they do not generalize to other label formats. 
\citet{Gaonkar20acl} as well as \citet{Wang21acl} learn vector representations for emotion classes from annotated text datasets to explicitly model their semantics and inter-relatedness. Yet again, these emotion embeddings (the class representations) do not generalize to other datasets and label formats. 
\citet{Han19taffc} propose a framework for learning a common embedding space as a means of joining information from different modalities in multimodal emotion data. While these embeddings generalize over different modalities (audio and video), they do not generalize across languages and label formats.
In summary, different from these studies, our emotion embeddings are not bound to any particular model architecture or dataset but instead generalize across domains and label formats, thus allowing to directly compare, say, English language items with BE5 ratings to Mandarin ones with VA ratings (see \tab\ref{tab:examples} vs.\ \fig\ref{fig:emotion-space}).

\paragraph{Coping with Incompatibility.}

In face of the variety of emotion formats,  \citet{Felbo17emnlp} present a transfer learning approach in which they pre-train a model with self-supervision to predict emojis in a large Twitter dataset, thus learning a representation that captures even subtle emotional nuances.  Similarly, multi-task learning can be used to fit a model on multiple datasets potentially having different label formats, thus resulting in shared hidden representations  \citep{Tafreshi18,Augenstein18naacl}. While representations learned with these approaches generalize across different label formats, they do not generalize across model architectures or language domains.

Cross-lingual approaches learn a common latent representation for different languages but these representations are often specific to only one pair of languages and do not generalize to other label formats \citep{Gao15cl,Abdalla17ijcnlp,Barnes18acl}.  
Similarly, recent work with Multilingual BERT \citep{Devlin19naacl} shows strong performance in cross-lingual zero-shot transfer \citep{Lamprinidis21wassa}, but samples from different languages still end up in different regions of the embedding space \citep{Pires19acl}. These approaches are also specific to a particular model architecture so that they do not naturally carry over to, e.g.,  single-word emotion prediction. 
Multi-modal approaches to emotion analysis show some similarity to our work, as they learn a common latent representation for several modalities which can be seen as separate domains  \citep{Zadeh17,Han19taffc,Poria19acl}. However, these representations are typically specific to a single dataset and are not meant to generalize further.

In a recent survey on text emotion datasets, \citet{Bostan18coling} point out naming inconsistencies between label formats. They build a joint resource that unifies twelve datasets under a common file format and annotation scheme. Annotations were unified based on the  semantic closeness of their class names (e.g., merging \textit{\enquote{happy}} and \enquote{\textit{Joy}}).  This approach is limited by its reliance on \textit{manually} crafted rules which are difficult to formulate, especially for numerical label formats.

In contrast, emotion representation mapping (or "label mapping")  aims at \textit{automatically} learning such conversion schemes between  formats from data (especially from ``double-annotated'' samples, such as the first two rows in \tab \ref{tab:examples}; \citealp{Stevenson07,Calvo13,Buechel18coling}). As the name suggests, label mapping operates exclusively on the gold ratings, without actually deriving representations for language items. It can, however, be used as a post-processor, converting the prediction of another model to an alternative label format (used as a baseline in \sec\ref{sec:setup}). Label mapping learns to transform \textit{one} format \textit{into another}, yet without establishing a more general representation. 
In a related study, \citet{DeBruyne22csl} indeed do learn a common representation for different label formats by applying variational autoencoders to multiple emotion lexicons. However, their method still only operates exclusively on the gold ratings without actually predicting labels based on words or texts.

In summary, while there are methods to learn common emotion representations across \textit{either} languages, linguistic domains, label formats, or model architectures, to the best of our knowledge, our proposal is the first to achieve all this simultaneously.

\section{Methods}

\begin{figure*}[tb]
	\centering
    \includegraphics{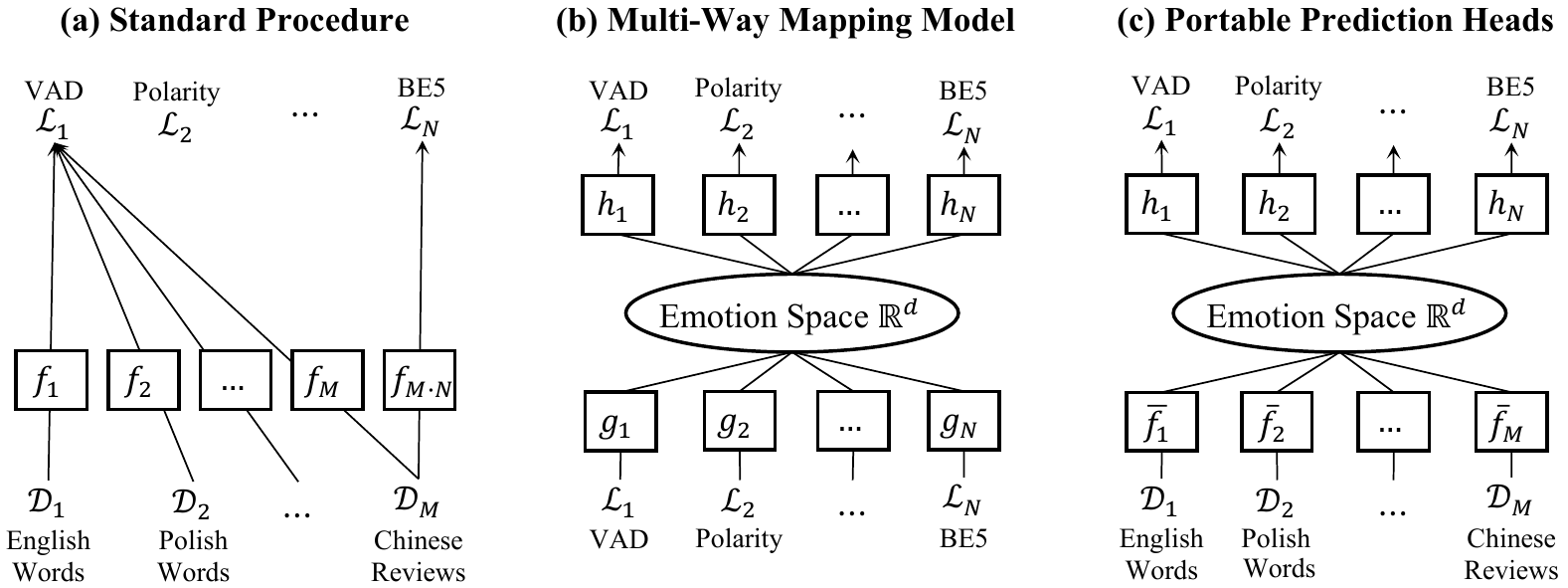}
	\caption{Overview of our methodology, illustrated by several choices of language domains and label formats.}
	\label{fig:architecture}
\end{figure*}

Let  $(X,Y)$ be a dataset with samples $X {\coloneqq} \{x_1, \dots x_n\}$ and labels $Y {\coloneqq} \{y_1, \dots, y_n\}$. The aim of emotion analysis is to find a model $f$ that best predicts $Y$ given $X$. Let us  assume that the samples $X$ are drawn from one of $M$ domains $\mathcal{D}_1, \dots, \mathcal{D}_M$ and the labels are drawn from one of $N$ label formats $\mathcal{L}_1, \dots, \mathcal{L}_N$. A domain refers to  the vocabulary or a particular register of a given language (word- and text-level prediction). A label format is a set of valid labels with reference to particular emotion constructs. For instance, the VAD format consists of vectors $(v,a,d)$ where the components $v,a,d$ refer to \textit{Valence}, \textit{Arousal}, and \textit{Dominance}, respectively, and are bound within a specified interval, e.g., $[1,9]$.

\subsection{Towards a Common Emotion Space}

\fig \ref{fig:architecture} provides an overview of our methodology. The na\"{i}ve approach to emotion analysis is to learn separate models for each language domain, $\mathcal{D}_1, \dots, \mathcal{D}_M$, and label format, $\mathcal{L}_1, \dots, \mathcal{L}_N$, resulting in a potentially very high number of relatively weak models in terms of the emotional nuances they can capture \textit{(a)}. The alternative we propose consists of two steps. First, we train a multi-way mapping that can translate between every pair of  label formats  $(\mathcal{L}_i, \mathcal{L}_j),\; i,j \in [1,N]$ via a shared intermediate representation layer, the common emotion space \textit{(b)}. In a second step, we adopt existing model architectures to embed samples from a given domain in the emotion space, while the format-specific top layers of said mapping model are now utilized as portable prediction heads. The emotion space then acts as a mediating ``interlingua'' which connects each language domain,  $\mathcal{D}_1, \dots, \mathcal{D}_M$, with each label format, $\mathcal{L}_1, \dots, \mathcal{L}_N$ \textit{(c)}.

\subsection{Prediction Head Training}
\label{sec:decoders}

A prediction head here refers to a function $h$ that maps from a Euclidean input space  $\mathbb{R}^d$ (the  "emotion space") to a label format $\mathcal{L}_j$. We give prediction heads a purposefully minimalist design that consists only of a single linear layer without bias term. Thus, a head $h$ predicts ratings $\hat{y}$ for an emotion embedding $x\in \mathbb{R}^d$ as $h(x) \coloneqq W x$, where $W$ is a weight matrix. The reason for this simple head design is to ensure that the affective information is more readily available in the emotion space. Alternatively, we can describe the weight matrix $W$ as a concatenation of row vectors $W_{i}$, where each emotion variable corresponds to exactly one row. Thus, as a positive side effect of the lightweight design, we can directly locate emotion variables within the emotion space by interpreting their respective coefficients $W_{i}$ as position vector  (see \fig \ref{fig:emotion-space}).

Our challenge is to train a collection of heads $h_1, \dots h_N$ such that all heads produce \textit{consistent} label outputs for a given emotion embedding from $\mathbb{R}^d$. For example, if the VAD head predicts a \textit{joyful} VAD label, then the BE5 head should also produce a congruent \textit{joyful} BE5 rating. In this sense, the prediction heads are \enquote{the heart and soul} of the emotion space: they define which affective state a region of the space corresponds to. 

To devise a suitable training scheme for the heads, we first need to elaborate on our understanding of "consistency" between differently formatted emotion labels. We argue that an obvious case of such consistency is found in datasets for emotion label mapping (see \sec \ref{sec:related}). A label mapping dataset consists of two sets of  labels following different formats $Y_1 {:=} \{y_{1,1},  y_{1,2},\dots y_{1,n}\}$ and $Y_{2} {:=} \{y_{2,1}, y_{2,2}, \dots y_{2,n}\}$, respectively. Typically, they are constructed by matching instances from independent annotation studies (e.g., the first two rows in \tab \ref{tab:examples}). Thus, we can think of the two sets of labels as \enquote{translational equivalents}, i.e., differently formatted emotion ratings, possibly capturing different affective nuances, yet still describing the same underlying expression of emotion in humans. 

The intuition behind our training scheme is to "fuse" multiple mapping models by forcing them to produce the same intermediate representation for both mapping directions.  This results in a multi-way mapping model with a shared representation layer in the middle (the common emotion space) followed by the prediction heads on top (\fig\ref{fig:architecture}\textit{b}).

\begin{figure}[b]
    \centering
    \includegraphics{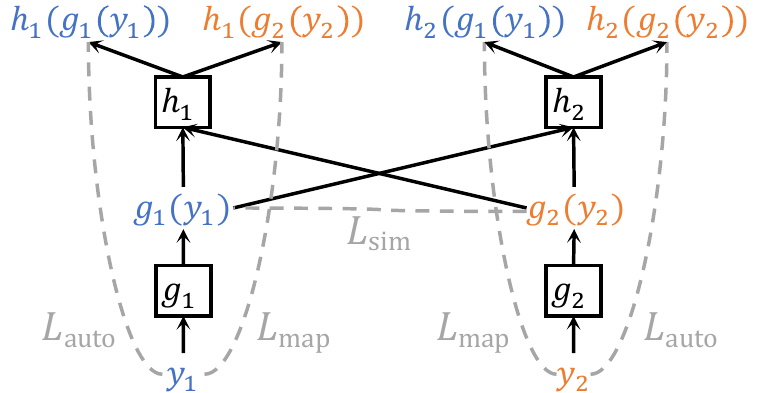}
    \caption{Training the Multi-Way Mapping Model.}
    \label{fig:head-training}
\end{figure}

In more detail (see also \fig \ref{fig:head-training} for an illustration of the following training procedure), let $(Y_1, Y_2)$ be a mapping dataset with a sample $(y_1, y_2)$. We introduce two new, auxiliary models $g_1, g_2$ that we call \textit{label encoders}. Label encoders embed input ratings in the emotion space $\mathbb{R}^d$ and  can be combined with the complementary prediction heads $h_2, h_1$ to form a mapping model  (the subscript here refers to the label format). That is $h_2(g_1(y_1))$ yields predictions for $y_2$ and $h_1(g_2(y_2))$ for $y_1$.

Our goal is to align both the intermediate representations, $g_1(y_1)$, $g_2(y_2)$  while also deriving accurate mapping predictions. Therefore, we propose the following three training objectives:
\begin{align*}
    L_\mathrm{map} &\coloneqq \mathcal{C}[y_1, h_1(g_2(y_2))] + \mathcal{C}[y_2 , h_2(g_1(y_1))]  \\
    L_\mathrm{auto} &\coloneqq \mathcal{C}[y_1 , h_1(g_1(y_1))] + \mathcal{C}[y_2, (h_2(g_2(y_2))] \\
    L_\mathrm{sim} &\coloneqq \mathcal{C}[g_1(y_1), g_2(y_2)]
\end{align*}
\noindent where $\mathcal{C}$ denotes the Mean-Squared-Error loss criterion. $L_\mathrm{map}$ is the  \textit{mapping loss} term where we compare true vs. predicted labels. The two summands represent the two mapping directions, assigning either of the two labels as the source, the other as the target format. The  \textit{autoencoder loss}, $L_\mathrm{auto}$, captures how well the model can reconstruct the original input label from the hidden emotion representation. It is meant to supplement the mapping loss. Lastly, the \textit{similarity loss}, $L_\mathrm{sim}$, directly assesses whether both input label formats end up with a similar intermediate representation. The \textit{total loss} for one instance, finally, is given by
\begin{equation*}
    L_\mathrm{total} := L_\mathrm{map} + L_\mathrm{auto} + L_\mathrm{sim}
\end{equation*}

In practice, we train a matching label encoder $g_1, \dots, g_N$ for each of our prediction heads $h_1, \dots, h_N$, thus covering all considered label formats $\mathcal{L}_1, \dots \mathcal{L}_N$. All label encoders and prediction heads are trained simultaneously on a collection of mapping datasets. This is done as a hierarchical sampling procedure, where we first sample one of the mapping datasets (which determines the encoder and the head to be optimized in this step), then a randomly selected instance. The total loss is computed in a batch-wise fashion and the encoder and head parameters are updated via standard gradient descent-based techniques (see Appendix \ref{appendix:pretraining} for details).
We use min-max scaling to normalize value ranges of the labels across  datasets: for VAD we choose the interval $[-1, 1]$ and for BE5 the interval $[0, 1]$, reflecting their respective bipolar (VAD) and unipolar (BE5) nature (see \tab \ref{tab:examples}).

\subsection{Prediction Head Deployment}
\label{sec:deployment}

\begin{figure}[t]
    \centering
    \includegraphics{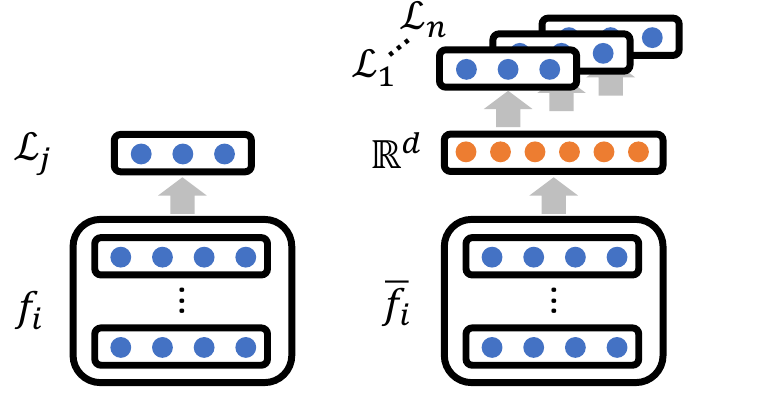}
    \caption{Schematic illustration of a base model before (left) and after (right) head deployment.}
    \label{fig:head-deployment}
\end{figure}

Following the training of the prediction heads $h_1, \dots, h_N$, deploying them on top of a base model architecture $f$ is relatively straightforward, resulting in a multi-headed model. The base model's output layer must be resized to the dimensionality of the emotion space $\mathbb{R}^d$ and any present nonlinearity (e.g, softmax or sigmoid activation) must be removed. This modified base model $\bar{f}$ is then optimized to produce emotion embeddings, the heads' input representation (see \fig \ref{fig:head-deployment}).

Head parameters are kept constant so that the base model is forced to optimize the representations it provides. Since the heads are specifically trained to treat emotion embeddings consistently, producing suitable representations for \textit{one} head is also likely to produce suitable representations for the remaining heads. Yet, to avoid overfitting the base model to a particular one (i.e., producing representations that are particularly favorable for one head, but much less so for every other), each model $\bar{f}_i$ is trained using multiple heads depending on the available data.

If \textit{multiple} datasets are available that match the domain of the base model \textit{and} use different label formats, we train the base model in a multi-task setup: We first draw one of the available datasets and then sample an instance $(x,y)$ from there. Next, we derive a prediction using the matching head $h_j$ as $\hat{y}  := h_j(\bar{f}_i(x))$, before computing the \textit{prediction loss}:
\begin{equation*}
L_\mathrm{pred} := \mathcal{C}[y, \hat{y}]    
\end{equation*}

If, on the other hand, only \textit{one} dataset is available which matches the domain of the base model $\bar{f}_i$, we complement the prediction loss with additional error signal using a newly proposed data augmentation technique. This method which we call \textit{emotion label augmentation} synthesizes an alternative label  $y^*:= h_k(g_j(y))$ for a given instance $(x,y)$ by taking advantage of the label encoder $g_j$ that was trained in the previous step. While $g_j$ translates the label $y$ to the  emotion space, the prediction head $h_k$ provides labels in a format different from $y$. Those artificial labels are then used in place of actual gold labels resulting in the \textit{data augmentation loss}
\begin{equation*}
L_\mathrm{aug} := \mathcal{C}[y^*, h_k(\bar{f}_i(x)]     
\end{equation*}
\noindent where the second argument to the loss criterion $\mathcal{C}$ denotes the model's prediction for the previously synthesized labels. Then,  $L_\mathrm{pred} + L_\mathrm{aug}$ yields the final loss.

\section{Experimental Setup}

The main idea behind our experimental setup is to compare a base model trained with the standard procedure against the same model with  portable prediction heads (PPH) attached (cf.\ \fig \ref{fig:architecture} \textit{(a)} vs.\ \textit{(c)}).  Our goal is to show that we obtain the same, if not better, results using PPH compared with the na\"{i}ve approach. 

This study design reflects two purposes. First, comparing the base model with the PPH architecture yields experimental data that allow to indirectly assess the quality of the learned emotion representations. Second, such a comparison may help find evidence that the performance of the PPH approach \textit{scales} with the employed base model---this would suggest that our method is likely to remain valuable even when today's state-of-the-art models are replaced by their successors. Importantly, we train only a single set of prediction heads. Thus, \textit{all} experimental results of the PPH condition are based on the \textit{same} underlying emotion space.

We distinguish two evaluation settings. In the first ("supervised") setting, train and test data come from (different parts of) the same dataset. Without PPH, we train one base model per dataset. Yet, with PPH, base models are shared across datasets of the same domain, whether or not their label formats agree. Consequently, the emotion space needs to store heterogeneous affective information in an easy-to-retrieve way (recall the "lightweight" head design; \sec\ref{sec:decoders}).  Thus, positive evaluation results would indicate that our method learns a particularly rich representation of emotion.  A practical advantage of PPH  lies in the reduction of total disk space utilized by the resulting model checkpoints. 

The second (\enquote{zero-shot}) setting assumes that only \textit{one} dataset per language is available, with one particular label format, but one would like to predict ratings in another format as well (e.g., imagine having a VA dataset for Mandarin but you are actually more interested in basic emotions for that language). Doing so with PPH is very simple---one only has to choose the desired head at inference time.  Yet, doing so with the base model \textit{per se} is simply impossible. To still be able to offer a quantitative comparison, we resort to an external label mapping component that translates the base model's output into the desired format. We emphasize that this is a very strong baseline due to the high accuracy of the label mapping approach, in general \cite{Buechel18coling}. In this case, the practical advantage of the PPH approach lies in its independence of (possibly unavailable) external post-processors.

\label{sec:setup}

\begin{table}[b!]
    \centering
    {\footnotesize{}
        \begin{tabular}{llrl}
\toprule
\textbf{ID} & \textbf{Vars} &   \textbf{Size} &         \textbf{Citation} \\
\midrule
en1 &   VAD & 1,034 & \scriptsize \citet{Bradley99anew} \\
en2 & BE5 & 1,034 & \scriptsize \citet{Stevenson07} \\
es1 &        VA &  14,031 &      \scriptsize \citet{Stadthagen17}            \\
es2 &     BE5 &  10,491 &        \scriptsize \citet{Stadthagen18}            \\
de1 &        VA &   2,902 &      \scriptsize \citet{Vo09}            \\
de2 &     BE5 &   1,958 &        \scriptsize \citet{Briesemeister11}            \\
pl1 &        VA &   2,902 &      \scriptsize \citet{Riegel15}            \\
pl2 &     BE5 &   2,902 &        \scriptsize \citet{Wierzba15}            \\
tr1 &        VA &   2,029 &      \scriptsize \citet{Kapucu18}            \\
tr2 &     BE5 &   2,029 &        \scriptsize \citet{Kapucu18}            \\
\bottomrule
\end{tabular}

    }
\caption{Word datasets. IDs contain the respective ISO 639-1 language code.
}

\label{tab:word-datasets}
\end{table}

\begin{table}[b!]
    \centering
    {\footnotesize{}
        \begin{tabular}{llrcp{2.7cm}}
\toprule
\textbf{ID} & \textbf{Vars} &   \textbf{Size} & \textbf{Lg} &  \textbf{Domain}  \\
\midrule
\textsc{AffT}& BE5 & 1,250 &  en  &  news headlines \\
\textsc{EmoB} &  VAD & 10,062  & en & genre-balanced   \\
\textsc{CVAT} & VA & 2,969 & zh & \mbox{mixed online domains}\\
\bottomrule
\end{tabular}

    }
    \caption{Overview of text datasets.}
    \label{tab:text-datasets}
    
\end{table} 

We conducted experiments on different word and text datasets.  For words, we collected ten datasets (cf.\ \tab \ref{tab:word-datasets}) covering five languages. These data are structured as illustrated in the top half of \tab \ref{tab:examples}. For text-level experiments we selected three corpora (cf.\  \tab{}\ref{tab:text-datasets}): Affective Text (\textsc{AffT}; \citealp{Strapparava07}),  \textsc{EmoBank} (\textsc{EmoB}; \citealp{Buechel17eacl}), and the Chinese Valence Arousal Texts (\textsc{CVAT}; \citealp{Yu16naacl}). For an illustration of the type and format of text-level data, see the bottom half in \tab \ref{tab:examples}. Since these datasets comprise real-valued annotations, we will use Pearson Correlation $r$ for measuring prediction quality. Datasets were partitioned into fixed train-dev-test splits with ratios ranging between 8-1-1 and 3-1-1; smaller datasets received larger dev and test shares. 

The selected data govern how to train a given base model with PPH (\sec\ref{sec:deployment}). Since, except for Mandarin, there are always two datasets available per domain, we train the models in the supervised setting using the multi-task approach (but use emotion label augmentation for CVAT). By contrast, in the zero-shot setting, we train a model on \textit{one}, yet  test on \textit{another} dataset. Thus, we rely on emotion label augmentation here (and have to exclude \textsc{CVAT} for a lack of a second Mandarin dataset). We emphasize that the zero-shot evaluation has very demanding data requirements: This setting not only requires two datasets of the \textit{same} language domain with \textit{different} label formats (which is already rare) but also additional data to fit mapping models for those particular label formats. To the best of our knowledge, \textsc{EmoBank} and \textsc{AffT} form the only suitable dataset pair on the text-level. At the word-level, such pairs are somewhat easier to get due to highly standardized data collection efforts for affective word norm datasets in psychology (see \S\ref{sec:related}). For this reason, we employ a larger number of 
word- than text-level datasets in our experiments.

Importantly, only the data requirements for \textit{evaluating} our approach in the zero-shot setting are hard to meet. Yet, \textit{inference} is much easier to provide. We would even argue that the reason why our method is so hard to evaluate is precisely what makes it so valuable. Take the Mandarin CVAT dataset, for example. It is annotated with \textit{Valence} and \textit{Arousal}, but there is, to our knowledge, no compatible Mandarin dataset with basic emotions (thus, CVAT is not used in the zero-shot setting). Our method allows to freely switch between output label formats at inference time without language constraints. That is, we can predict BE5 ratings in Chinese even though there is no such training data.

In terms of base models, we used the Feed-Forward Network developed by \citet{Buechel18naacl} for the word datasets. This model predicts emotion ratings based on pre-trained embedding vectors (taken from \citealp{Grave18lrec}). For text datasets, we chose the $\textsc{Bert}_\mathrm{base}$ transformer model by \citet{Devlin19naacl} using the implementation and pre-trained weights by \citet{Wolf20arxiv}. Both (word and text) base models use identical hyperparameter settings with or without PPH extension. For the word model, we copied the settings of the authors, whereas text model hyperparameters were tuned manually for the base model \textit{without} PPH. 

We derived training data for the prediction heads (label mapping datasets) by combining the ratings of the word datasets $en1$ and $en2$.  We used the label mapping model from \citet{Buechel18coling} as auxiliary label encoders. The dimensionality of the emotion space was set to 100. The label mapping models used as external post-processors in the zero-shot setting were also based on \citet{Buechel18coling} and were trained on the same data as the label encoders. Further details beneficial for reproducibility are given in Appendix \ref{app:setup}.

\section{Results} 
\label{sec:results}

Our main experimental results are summarized in Tables \ref{tab:results-supervised-word} to \ref{tab:results-zeroshot-text}. For conciseness, correlation values are averaged over all target variables per dataset. Per-variable results are given in Appendix \ref{app:per-variable-results}.

Looking at the word datasets in the supervised setup (\tab \ref{tab:results-supervised-word}), we find that  attaching portable prediction heads (PPH) not only retains, but often enough slightly increases the performance  of the FFN base model ($p{=}.008$; two-sided Wilcoxon signed-rank test based on per-dataset results). Since we trained only one base model with PPH per language (but two without PPH), our data suggest that the emotion representations learned with PPH can easily hold affective information from different label formats at the same time. Moreover, PPH here offers the practical benefit of reducing the total disk space used by the resulting model checkpoints due to the smaller number of trained base models. Experiments on the text datasets using BERT as base model show results in line with these findings (see \tab \ref{tab:results-supervised-text}). 

\begin{table}[b!]
    \centering
    \footnotesize
    \begin{tabular}{|l|l|r|l|r|}
    \hline
       & \multicolumn{2}{c|}{\textbf{Base Model (FFN)}} & \multicolumn{2}{c|}{\textbf{Base Model + PPH}} \\
    Test Data & Train Data & $r$ & Train Data & $r$ \\
    \hline
    en1\texttt{(VAD)} & en1\texttt{(VAD)} & .818 &  en1+en2 & .824 \\
    en2\texttt{(BE5)} & en2\texttt{(BE5)} & .898 & en1+en2 & .898 \\
    es1\texttt{(VA)} & es1\texttt{(VA)} & .820 & es1+es2 & .833 \\
    es2\texttt{(BE5)}  & es2\texttt{(BE5)} & .789 & es1+es2 & .820 \\
    de1\texttt{(VA)} & de1\texttt{(VA)} & .822 & de1+de2 & .836 \\
    de2\texttt{(BE5)} & de2\texttt{(BE5)} & .754 & de1+de2 & .748 \\
    pl1\texttt{(VA)} & pl1\texttt{(VA)} & .794 & pl1+pl2 & .835 \\
    pl2\texttt{(BE5)} & pl2\texttt{(BE5)} & .814 & pl1+pl2 & .845 \\
    tr1\texttt{(VA)} & tr1\texttt{(VA)} & .567 & tr1+tr2 & .575 \\
    tr2\texttt{(BE5)} & tr2\texttt{(BE5)} & .607 & tr1+tr2 & .614 \\
    \hline
    Mean & \multicolumn{2}{r|}{.768} & \multicolumn{2}{r|}{.783} \\
    Disk Use & \multicolumn{2}{r|}{4.33 MB} & \multicolumn{2}{r|}{2.52 MB} \\ 
    \hline
\end{tabular}
    \caption{Word-level results of supervised setting.}
    \label{tab:results-supervised-word}
\end{table}

\begin{table}[b!]
    \centering
    \footnotesize
    \begin{tabular}{|l|l|r|l|r|}
    \hline
       & \multicolumn{2}{c|}{\textbf{Base Model (BERT)}} & \multicolumn{2}{c|}{\textbf{Base Model + PPH}} \\
    Test Data & Train Data & $r$ & Train Data & $r$ \\
    \hline
    EmoB & EmoB & .630 & EmoB+AffT & .619 \\
    AffT & AffT & .746 & EmoB+AffT & .755 \\
    CVAT & CVAT & .737 & CVAT & .748 \\
    \hline
    Mean & \multicolumn{2}{r|}{ .704} & \multicolumn{2}{r|}{ .707} \\
    {Disk Use} & \multicolumn{2}{r|}{1.25 GB} & \multicolumn{2}{r|}{0.81 GB} \\ 
    \hline
\end{tabular}
    \caption{Text-level results of supervised setting.}
    \label{tab:results-supervised-text}
\end{table}

In the zero-shot setup, models are tested on datasets with label formats different from the training phase (e.g., $en1$ and $en2$).  
On the word datasets, using PPH shows small improvements in comparison with the base model as is ($p{=}.003$; \tab \ref{tab:results-zeroshot-word}), again suggesting that the learned emotion representations generalize robustly across label formats. Importantly, the base model is only capable of producing this label format \textit{at all} because we equip it with a label mapping post-processor. While this procedure is very accurate (indeed, it constitutes a very strong baseline), it depends on an external component that may or may not be available for the desired mapping direction (the source and the target label format). In contrast, the zero-shot capability is \textit{innate} to (``built-in'') the PPH approach. While we need only one prediction head per label format, the number of required mapping components for the base model grows on a quadratic scale with the number of considered formats. Again, text-level experiments show consistent results with word-level ones (\tab{} \ref{tab:results-zeroshot-text}).

\begin{table}[t]
    \centering
    \footnotesize
    \begin{tabular}{|l|l|r|l|r|}
    \hline
        & \multicolumn{2}{c|}{\textbf{Base Model (FFN)}} & \multicolumn{2}{c|}{\textbf{Base Model + PPH}} \\
    Test Data & Train Data & $r$ & Train Data & $r$ \\
    \hline
    en1\texttt{(VAD)} & en2\texttt{(BE5)} & .801 & en2 & .810 \\ 
    en2\texttt{(BE5)} & en1\texttt{(VAD)} & .834 & en1 & .839 \\
    es1\texttt{(VA)} & es2\texttt{(BE5)} & .720 & es2 & .723 \\
    es2\texttt{(BE5)} & es1\texttt{(VA)} & .777 & es1 & .792 \\
    de1\texttt{(VA)} & de2\texttt{(BE5)} & .681 & de2 & .684 \\
    de2\texttt{(BE5)} & de1\texttt{(VA)} & .637 & de1 & .641 \\
    pl1\texttt{(VA)} & pl2\texttt{(BE5)} & .812 & pl2 & .812 \\
    pl2\texttt{(BE5)} & pl1\texttt{(VA)} & .787 & pl1 & .807 \\
    tr1\texttt{(VA)} & tr2\texttt{(BE5)} & .538 & tr2 & .563 \\
    tr2\texttt{(BE5)} & tr1\texttt{(VA)} & .550 & tr1 & .554 \\
    \hline
    Mean & \multicolumn{2}{r|}{.714} & \multicolumn{2}{r|}{.723} \\
    Method  & \multicolumn{2}{r|}{ext. post-processor} & \multicolumn{2}{r|}{built-in} \\
    \hline
\end{tabular}
    \caption{Word-level results of zero-shot setting.}
    \label{tab:results-zeroshot-word}
\end{table}

\begin{table}[t]
    \centering
    \footnotesize
    \begin{tabular}{|l|l|r|l|r|}
    \hline
      & \multicolumn{2}{c|}{\textbf{Base Model (BERT)}} & \multicolumn{2}{c|}{\textbf{Base Model + PPH}} \\
    Test Data & Train Data & $r$ & Train Data & $r$ \\
    \hline
    EmoB & AffT & .385 & AffT & .407 \\
    AffT & EmoB & .584 & EmoB & .582 \\
    \hline
    Mean & \multicolumn{2}{r|}{ .485} & \multicolumn{2}{r|}{ .495} \\
     Method & \multicolumn{2}{r|}{ext. post-processor} & \multicolumn{2}{r|}{built-in} \\
    \hline
\end{tabular}
    \caption{Text-level results of zero-shot setting.}
    \label{tab:results-zeroshot-text}
\end{table}

One may object that the reduction of memory footprint shown in Tables \ref{tab:results-supervised-word} and \ref{tab:results-supervised-text} can also be achieved by traditional multi-task learning (i.e., attaching multiple heads to the base model, training it on two datasets, at once). Likewise, as Tables \ref{tab:results-zeroshot-word} and \ref{tab:results-zeroshot-text} indicate, the zero-shot capabilities offered by PPH can, in principle, be provided by additional label mapping components. However, PPH offers a much more elegant solution to combine the advantages of multi-task learning and label mapping without calling for additional (language) resources. Most importantly though, PPH is unique in its ability to embed samples from such heterogeneous datasets in a common representation space---a trait that may offer a general solution to studying emotion across languages, cultures, and individually preferred psychological theory.

\section{Visualization of the Emotion Space} 
\label{sec:visualization}
To gain first insights into the structure of our learned emotion space, we submitted the weight vectors of the emotion variables to principal component analysis (PCA; recall from \sec \ref{sec:decoders} that each row in a head's weights matrix $W$ corresponds to exactly one variable). Further, we derived emotion embeddings for the samples in \tab \ref{tab:examples} using the PPH-extended models evaluated in the last section. Applying the same PCA transformation to the embedding vectors, we co-locate the samples next to the emotion variables. The results (for the first three PCs) are displayed in \fig \ref{fig:emotion-space}.  As can be seen, the relative positioning of the samples and variables shows high face validity---samples associated with similar feelings appear close to each other as well as to their akin variable. 
Appendix \ref{app:analysis} provides additional analyses of the learned embedding space (focusing more deeply on the emotional interpretation of the PC axes and the distribution of emotion embeddings across languages) that further support this positive impression.

\section{Conclusions \& Future Work}
\label{sec:conclusion}

We presented a method for learning a common representation space for the emotional loading of heterogeneous language items. While previous work successfully unified \textit{some} sources' heterogeneity, our emotion embeddings are the first to \textit{comprehensively generalize} over arbitrarily disparate language domains, label formats, and distinct neural network architectures. Our technique is based on a collection of \textit{portable prediction heads} that can be attached to existing state-of-the-art models. Consequently, a model learns to \textit{embed} language items in the common learned emotion space and thus to predict a wider range of emotional meaning facets, yet without sacrificing any predictive power as our experiments on 13 datasets (6 languages) indicate.

Since the resulting emotion representations both generalize across various use cases \textit{and} evidently capture a rich set of affective nuances, we consider this work particularly useful for downstream applications. Thus, future work may build on a concept of \textit{emotion similarity} to, e.g., cluster diverse language items by their associated feeling, retrieve words that evoke emotions similar to a query, or compare the affective meaning of phrases and concepts across cultures. 

\section*{Acknowledgments}

We would like to thank the anonymous reviewers for their helpful suggestions and comments, and Tinghui Duan, doctoral student at the \textsc{Julie Lab}, for assisting us with the Mandarin gold data.

\bibliography{literature}
\bibliographystyle{stylesheets/emnlp2021/acl_natbib}
\clearpage

\appendix

\section{Algorithmic Details for Training the Multi-Way Mapping Model}
\label{appendix:pretraining}

\begin{algorithm*}
	\caption{Training the Multi-Way Mapping Model}
	\label{alg:pretraining}
	\begin{algorithmic}[1]

\STATE $(Y_{1,1}, Y_{1,2}), (Y_{2,1}, Y_{2,2}), \dots (Y_{n,1}, Y_{n,2}) \leftarrow$ Mapping datasets used for training 
\STATE $g_{1,1}, h_{1,1}, g_{1,2}, h_{1,2}, \dots, g_{n,1}, h_{n,1}, g_{n,2}, h_{n,2}\leftarrow$  randomly initialized label encoders and prediction heads \textsuperscript{\dag}
\STATE $n_\mathrm{steps} \leftarrow$ total number of training steps
\FORALL{$i_\mathrm{step}$ in $1, \dots, n_\mathrm{steps}$}
\STATE $(Y_{i,1}, Y_{i,2}) \leftarrow$ randomly sample a mapping dataset \label{algline:start-sampling}
\STATE $(y_1, y_2) \leftarrow$  randomly sample a batch s.t. $y_1 \subset Y_{i,1}$ and $y_{2} \subset Y_{i,2}$ with identical indices \label{algline:stop-sampling}
\STATE $(e_1, e_2) \leftarrow (g_{i,1}(y_1), g_{i,2}(y_2))$ \label{algline:encoding}
\STATE $\hat{y}_{1,1} \leftarrow h_{i,1}(e_1)$ \label{algline:start-decoding}
\STATE $\hat{y}_{1,2} \leftarrow h_{i,2}(e_1)$
\STATE $\hat{y}_{2,1} \leftarrow h_{i,1}(e_2)$
\STATE $\hat{y}_{2,2} \leftarrow h_{i,2}(e_2)$  \label{algline:stop-decoding}
\STATE $L_\mathrm{map} \leftarrow \mathcal{C}(y_1,\hat{y}_{2,1}) + \mathcal{C}(y_2, \hat{y}_{1,2}) $ \textsuperscript{\ddag}\label{algline:pred} 
\STATE $L_\mathrm{auto} \leftarrow \mathcal{C}(y_1,\hat{y}_{1,1}) + \mathcal{C}(y_2, \hat{y}_{2,2})$ \label{algline:auto} 
\STATE $L_\mathrm{sim} \leftarrow \mathcal{C}(e_1,e_2)$ \label{algline:sim} 
\STATE $L_\mathrm{total} \leftarrow L_\mathrm{map} + L_\mathrm{auto} + L_\mathrm{sim}$
\STATE compute $\nabla L_\mathrm{total}$ and update weights
\ENDFOR
\par\noindent\rule{.2\columnwidth}{0.4pt}
\par\noindent {
\textsuperscript{\dag} If two sets of labels $Y_{a,b}, Y_{c,d}$ follow the same label format, then they use the same label encoders (i.e, $g_{a,b} = g_{c,d}$) and prediction heads ($h_{a,b} = h_{c,d}$).\\
\vspace{0.5em}\textsuperscript{\ddag} $\mathcal{C}$ denotes Mean-Squared-Error Loss.
}
\end{algorithmic}

\end{algorithm*}

The intuition behind Algorithm \ref{alg:pretraining} is as follows: We simultaneously train multiple label encoders and prediction heads on several mapping datasets using three distinct objective functions. First, of course, we consider the quality of the label mapping (\textit{mapping loss}; line \ref{algline:pred}). Second, we propose an \textit{autoencoder loss} (line \ref{algline:auto})  where the model must learn to reconstruct the original input from the emotion embedding. Third, we propose an \textit{embedding similarity loss} (line \ref{algline:sim}) which enforces the similarity of the hidden representation of both formats for a given instance since they supposedly describe the same emotion. Our training loop starts by first sampling one of the mapping datasets and then a batch from the chosen dataset (lines \ref{algline:start-sampling}--\ref{algline:stop-sampling}).  To compute the loss efficiently, we first cache the encoded representations of both label formats (line \ref{algline:encoding}) before applying all relevant prediction heads (lines \ref{algline:start-decoding}--\ref{algline:stop-decoding}).

\section{Per-Variable Results}
\label{app:per-variable-results}

For readability reasons, the experimental results reported in \sec\ref{sec:results} only give the average performance score over all emotional target variables for a given dataset. To complement this, the full set of per-variable results are given in \tab\ref{tab:full-results}.

\begin{table*}[ht]
    \centering
    \footnotesize
    \setlength{\tabcolsep}{4pt}
    \begin{tabular}{lllllrrrrrrrrr}
\toprule
     &     &          &         &       &    Val &    Aro &    Dom &    Joy &    Ang &    Sad &   Fea &    Dis &   Mean \\
Level & Test & Setting & Model & Train &        &        &        &        &        &        &        &        &        \\
\midrule
word & en1 & supervised & FFN &   en1    &  .920 &  .704 &  .829 &    --- &    --- &    --- &    --- &    --- &  .818 \\
     &     &          & FFN+PPH &   en1+en2    &  .936 &  .700 &  .836 &    --- &    --- &    --- &    --- &    --- &  .824 \\
     &     & zeroshot & FFN &   en2    &  .932 &  .664 &  .808 &    --- &    --- &    --- &    --- &    --- &  .801 \\
     &     &          & FFN+PPH &   en2    &  .927 &  .701 &  .802 &    --- &    --- &    --- &    --- &    --- &  .810 \\
     & en2 & supervised & FFN &   en2    &    --- &    --- &    --- &  .929 &  .900 &  .898 &  .890 &  .873 &  .898 \\
     &     &          & FFN+PPH &  en1+en2    &    --- &    --- &    --- &  .936 &  .890 &  .895 &  .901 &  .869 &  .898 \\
     &     & zeroshot & FFN &   en1    &    --- &    --- &    --- &  .918 &  .822 &  .805 &  .864 &  .759 &  .834 \\
     &     &          & FFN+PPH &  en1     &    --- &    --- &    --- &  .914 &  .835 &  .850 &  .843 &  .751 &  .839 \\
     & es1 & supervised & FFN &   es2    &  .848 &  .792 &    --- &    --- &    --- &    --- &    --- &    --- &  .820 \\
     &     &          & FFN+PPH &  es1+es2     &  .870 &  .795 &    --- &    --- &    --- &    --- &    --- &    --- &  .833 \\
     &     & zeroshot & FFN &   es2    &  .873 &  .567 &    --- &    --- &    --- &    --- &    --- &    --- &  .720 \\
     &     &          & FFN+PPH &   es2    &  .872 &  .575 &    --- &    --- &    --- &    --- &    --- &    --- &  .723 \\
     & es2 & supervised & FFN &   es2    &    --- &    --- &    --- &  .768 &  .793 &  .834 &  .803 &  .745 &  .789 \\
     &     &          & FFN+PPH &  es1+es2     &    --- &    --- &    --- &  .817 &  .832 &  .857 &  .838 &  .754 &  .820 \\
     &     & zeroshot & FFN &   es1    &    --- &    --- &    --- &  .808 &  .795 &  .823 &  .775 &  .685 &  .777 \\
     &     &          & FFN+PPH &  es2     &    --- &    --- &    --- &  .811 &  .805 &  .839 &  .810 &  .695 &  .792 \\
     & de1 & supervised & FFN &   de1    &  .867 &  .776 &    --- &    --- &    --- &    --- &    --- &    --- &  .822 \\
     &     &          & FFN+PPH &  de1+de2     &  .892 &  .780 &    --- &    --- &    --- &    --- &    --- &    --- &  .836 \\
     &     & zeroshot & FFN &    de2   &  .832 &  .530 &    --- &    --- &    --- &    --- &    --- &    --- &  .681 \\
     &     &          & FFN+PPH &  de2     &  .836 &  .532 &    --- &    --- &    --- &    --- &    --- &    --- &  .684 \\
     & de2 & supervised & FFN &   de2    &    --- &    --- &    --- &  .812 &  .766 &  .738 &  .798 &  .653 &  .754 \\
     &     &          & FFN+PPH &  de1+de2     &    --- &    --- &    --- &  .842 &  .788 &  .655 &  .795 &  .662 &  .748 \\
     &     & zeroshot & FFN &  de1     &    --- &    --- &    --- &  .824 &  .717 &  .500 &  .733 &  .411 &  .637 \\
     &     &          & FFN+PPH &   de1    &    --- &    --- &    --- &  .824 &  .720 &  .489 &  .749 &  .424 &  .641 \\
     & pl1 & supervised & FFN &  pl1     &  .852 &  .735 &    --- &    --- &    --- &    --- &    --- &    --- &  .794 \\
     &     &          & FFN+PPH &  pl1+pl2     &  .907 &  .764 &    --- &    --- &    --- &    --- &    --- &    --- &  .835 \\
     &     & zeroshot & FFN &   pl2    &  .919 &  .705 &    --- &    --- &    --- &    --- &    --- &    --- &  .812 \\
     &     &          & FFN+PPH &  pl2     &  .918 &  .707 &    --- &    --- &    --- &    --- &    --- &    --- &  .812 \\
     & pl2 & supervised & FFN & pl2  &    --- &    --- &    --- &  .819 &  .807 &  .815 &  .810 &  .821 &  .814 \\
     &     &          & FFN+PPH &  pl1+pl2     &    --- &    --- &    --- &  .897 &  .835 &  .820 &  .826 &  .846 &  .845 \\
     &     & zeroshot & FFN &  pl1     &    --- &    --- &    --- &  .877 &  .786 &  .749 &  .763 &  .761 &  .787 \\
     &     &          & FFN+PPH &  pl1     &    --- &    --- &    --- &  .893 &  .798 &  .777 &  .779 &  .789 &  .807 \\
     & tr1 & supervised & FFN &   tr1    &  .556 &  .577 &    --- &    --- &    --- &    --- &    --- &    --- &  .567 \\
     &     &          & FFN+PPH &  tr1+tr2     &  .571 &  .579 &    --- &    --- &    --- &    --- &    --- &    --- &  .575 \\
     &     & zeroshot & FFN &  tr2     &  .561 &  .514 &    --- &    --- &    --- &    --- &    --- &    --- &  .538 \\
     &     &          & FFN+PPH &   tr2    &  .576 &  .549 &    --- &    --- &    --- &    --- &    --- &    --- &  .563 \\
     & tr2 & supervised & FFN &    tr1   &    --- &    --- &    --- &  .607 &  .603 &  .628 &  .627 &  .568 &  .607 \\
     &     &          & FFN+PPH &   tr1+tr2    &    --- &    --- &    --- &  .611 &  .608 &  .628 &  .634 &  .589 &  .614 \\
     &     & zeroshot & FFN &   tr1    &    --- &    --- &    --- &  .547 &  .566 &  .563 &  .579 &  .495 &  .550 \\
     &     &          & FFN+PPH &   tr1    &    --- &    --- &    --- &  .583 &  .533 &  .575 &  .588 &  .488 &  .554 \\
     \midrule
text & EmoB & supervised & BERT &   EmoB    &  .801 &  .562 &  .527 &    --- &    --- &    --- &    --- &    --- &  .630 \\
     &      &            & BERT+PPH &  EmoB+AffT     &  .798 &  .550 &  .509 &    --- &    --- &    --- &    --- &    --- &  .619 \\
     &      & zeroshot & BERT &   AffT    &  .660 &  .200 &  .295 &    --- &    --- &    --- &    --- &    --- &  .385 \\
     &      &            & BERT+PPH &   AffT    &  .686 &  .238 &  .297 &    --- &    --- &    --- &    --- &    --- &  .407 \\
     & AffT & supervised & BERT &  AffT     &    --- &    --- &    --- &  .730 &  .634 &  .818 &  .836 &  .712 &  .746 \\
     &      &            & BERT+PPH &  EmoB+AffT     &    --- &    --- &    --- &  .776 &  .659 &  .823 &  .841 &  .675 &  .755 \\
     &      & zeroshot & BERT &   EmoB    &    --- &    --- &    --- &  .727 &  .485 &  .727 &  .689 &  .290 &  .584 \\
     &      &            & BERT+PPH &  EmoB     &    --- &    --- &    --- &  .724 &  .491 &  .736 &  .704 &  .255 &  .582 \\
     & CVAT & supervised & BERT &   CVAT    &  .878 &  .596 &    --- &    --- &    --- &    --- &    --- &    --- &  .737 \\
     &      &            & BERT+PPH &  CVAT     &  .878 &  .617 &    --- &    --- &    --- &    --- &    --- &    --- &  .748 \\
\bottomrule
\end{tabular}

    \caption{Full experimental results per dataset and target variable in Pearson's $r$. "Mean" column corresponds to data given in  Tabs. \ref{tab:results-supervised-word}, \ref{tab:results-supervised-text}, \ref{tab:results-zeroshot-word}, and \ref{tab:results-zeroshot-text}.}
    \label{tab:full-results}
\end{table*}

\FloatBarrier

\section{Further Analysis of the Emotion Space}
\label{app:analysis}

\begin{figure}[t]
    \centering
    \includegraphics[width=.49\textwidth]{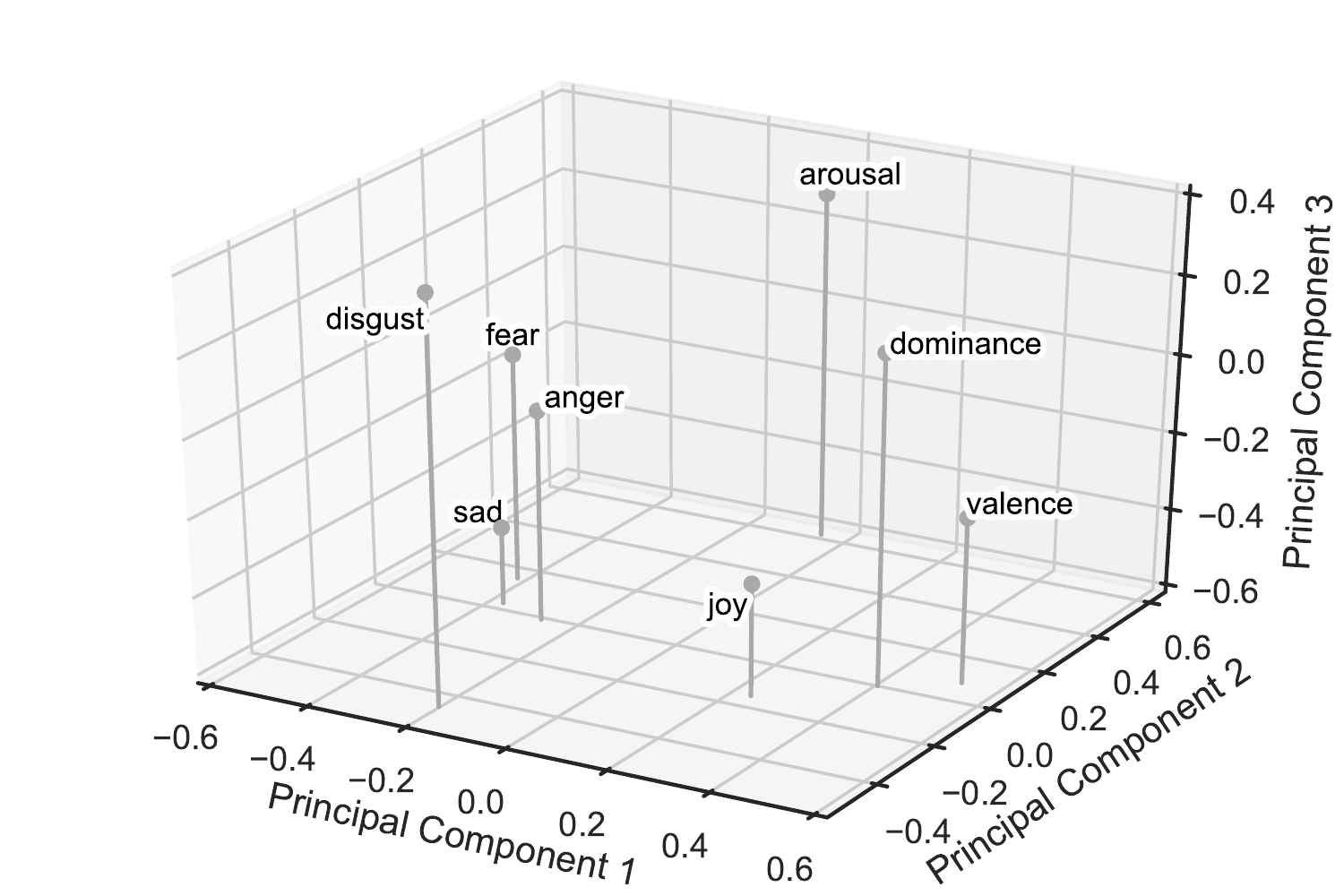}
    \caption{Position of emotion variables in PCA space.}
    \label{fig:variables}
\end{figure}

Building on the PCA transformation described in \sec \ref{sec:visualization}, we illustrate the position of \textit{all} emotion variables in \fig \ref{fig:variables}.

Within the first three principal components, two major groups can be visually discerned: the negative basic emotions of \textit{Sadness}, \textit{Fear}, and \textit{Anger} forming the first group, and \textit{Joy} and the two affective dimensions of \textit{Valence} and \textit{Dominance} forming the second. Intuitively speaking, this stands to reason, as \textit{Valence} and \textit{Dominance} typically show a very high positive correlation in annotation studies. The same holds for \textit{Valence} and \textit{Joy}.  Likewise, \textit{Sadness}, \textit{Fear}, and \textit{Anger} usually correlate positively with each other.  Yet, between these groups of variables, studies show a negative correlation (cf. studies listed in \tab \ref{tab:word-datasets}). Interestingly, these observations indicate that the first principal component of the emotion space may represent a \textit{Polarity} axis. 

The remaining two variables, \textit{Disgust} and \textit{Arousal}, position themselves relatively far from the aforementioned groups and opposite of each other in the second principal component.  While it is less obvious what this component represents, it is worth noting that both \textit{Arousal} and \textit{Disgust} generalize poorly across label formats. That is, while \textit{Joy}, \textit{Anger}, \textit{Sadness}, and \textit{Fear} are relatively easy to predict from VAD ratings in a label mapping experiment, and, likewise,  \textit{Valence} and \textit{Dominance} can well be estimated from BE5 ratings, the variables of \textit{Arousal} and \textit{Disgust} seem to carry information more specific to their respective label format \citep{Buechel18coling}. In the light of these observations, it may not come as a surprise that these variables receive positions that demarcate them clearly from the remaining ones. 

The third principal component seems to be linked to the intensity or action potential of a feeling. Here, \textit{Arousal}, \textit{Dominance}, and \textit{Disgust} and, less pronounced, \textit{Fear} and \textit{Anger} score highly, while \textit{Sadness} and \textit{Joy} receive comparatively low values.  

\begin{figure*}[tb]
    \centering
    \includegraphics[width=\textwidth]{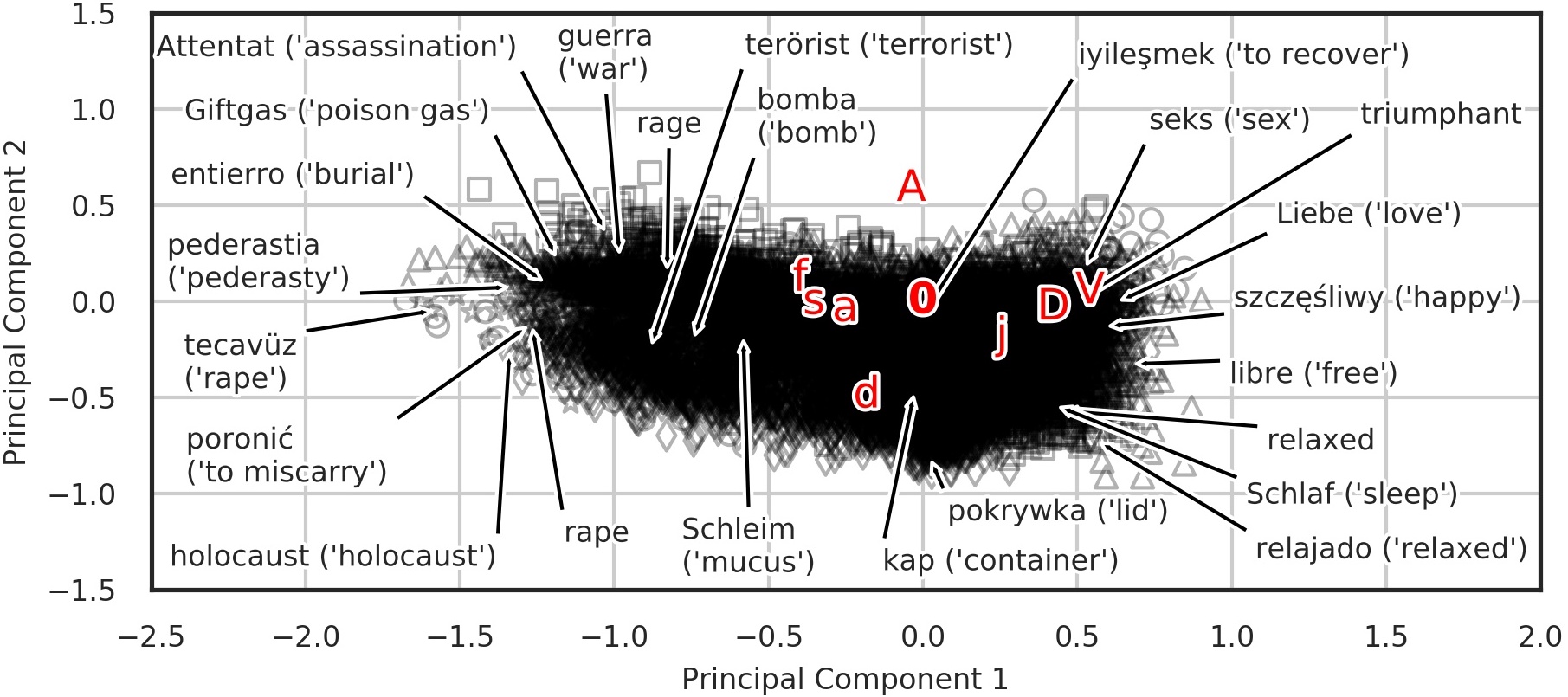}
    \vspace*{-1.5em}
   	\caption{
   		Position of the emotion variables \textbf{V}\textit{alence}, \textbf{A}\textit{rousal}, \textbf{D}\textit{ominance} and \textbf{j}\textit{oy}, \textbf{a}\textit{nger}, \textbf{s}\textit{adness}, \textbf{f}\textit{ear}, and \textbf{d}\textit{isgust}  in the learned emotion space $\mathbb{R}^d$ (first two PCA dimensions; origin marked with \enquote{\textbf{0}}) together with entries from English ({\footnotesize $\bigcirc$}), Spanish ($\bigtriangleup$), German ($\square$), Polish ($\diamondsuit$), and Turkish (\FiveStarOpen) word datasets, as well as highest and lowest \textit{Valence} and \textit{Arousal} word, and highest \textit{Disgust} word per language (arrows).
    \label{fig:lexicon-emotion-space}
} 
\end{figure*}

Next, we examine whether the learned representations are sufficiently language-agnostic, i.e., that samples with similar emotional load receive similar embeddings independent of their language domain. We derived emotion embeddings for all entries in all of our word datasets (cf.\ \tab\ref{tab:word-datasets})  using the base models with portable prediction heads from the "supervised" setting of our main experiments. Again building on the previously established PCA transformation, we plotted the position of these multilingual samples in 2D (see \fig \ref{fig:lexicon-emotion-space}). 

 It is noteworthy that entries in our emotion space seem to form clusters according to their affective meaning and not within their dataset or language. As a result, items from different languages overlap so heavily that their respective markers (\raisebox{0.1em}{\footnotesize $\bigcirc$}, $\bigtriangleup$,$\square$,$\diamondsuit$, and \raisebox{-0.1em}{\FiveStarOpen}) become hard to differentiate.

Furthermore, we selected the highest- and lowest-rated words for \textit{Valence} and \textit{Arousal} and the highest-rated word for \textit{Disgust} in each language. We locate these words in the PCA space and give translations for non-English entries. As can be seen, their position shows high face validity relative to each other and the emotion variables,  supporting our claim that the learned emotion space is indeed language-independent. 

We emphasize that monolingual, rather than crosslingual, word embeddings were used and that samples from each language were embedded using a separate base model. Hence, the observed alignment of words in PCA space may safely be attributed to our proposed training scheme using portable prediction heads.

\section{Further Details for Reproducibility}
\label{app:setup}

\subsection{Description of Computing Infrastructure}
All experiments were conducted on a single machine with a Debian 4  operating system. The hardware specifications are as follows: 
\begin{itemize}
\item 1 GeForce GTX 1080 with 8 GB graphics memory
\item 1 Intel i7 CPU with 3.60 GHz
\item 64 GB RAM
\end{itemize}

\subsection{Runtime of the Experiments}

Training the multi-way mapping model takes about one minute. Training time for the base models varies depending on the dataset. In the following, we report training and inference times for the \textit{largest} dataset per condition, respectively, describing an upper bound of the time requirements.

Regarding the word models, it takes about ten minutes to train a base model without portable prediction heads (PPH) and about 15 minutes to train one with PPH. Since the latter base model replaces two of the former ones in our experiments, the overall training time is reduced by using PPH. Training a word model with emotion label augmentation (the alternative technique for fitting a model with PPH) takes 10 minutes, about as long as training it without PPH. Inference is  completed in 1.5 minutes in either case. However, most of that time is needed for loading the language-specific word embeddings. Once this task is done, actually computing the predictions takes only about one second.

Regarding the text models, a baseline model without PPH is trained in about 15 minutes. This number increases with PPH to 30 minutes using the multi-task approach (but again, one PPH model replaces two of the baseline models). In line with the runtime results of the word models, training the text base model with emotion label augmentation takes 15 minutes, about as long as training it without PPH. In either case, inference is completed in well under a minute. 

\subsection{Number of Parameters in Each Model}

The number of parameters per model is given in \tab\ref{tab:parameters}.

\begin{table}[ht]
    \centering
    \begin{tabular}{l|r}
        Model (Component) & No. Parameters \\
        \hline
        Portable Prediction Heads & 0.8K \\ 
        Label Encoders (per format) & 18.8K \\ 
        Label Encoders (in total) & 53.4K \\ 
        Word-Level FFN (per model) & 110.6K \\ 
        BERT\textsubscript{base} (per model) & 110.0M \\ 
    \end{tabular}
\caption{Number of parameters in each model.}
\label{tab:parameters}
\end{table}

\subsection{Validation Performance}

\begin{table}[t]
\centering
\footnotesize
\begin{tabular}{|l|l|r|l|r|}
    \hline
       & \multicolumn{2}{c|}{\textbf{Base Model (FFN)}} & \multicolumn{2}{c|}{\textbf{Base Model + PPH}} \\
    Test Data & Train Data & $r$ & Train Data & $r$ \\
    \hline
    en1\texttt{(VAD)} & en1\texttt{(VAD)} & .800 &  en1+en2 & .806 \\
    en2\texttt{(BE5)} & en2\texttt{(BE5)} & .876 & en1+en2 & .877 \\
    es1\texttt{(VA)} & es1\texttt{(BE5)} & .832 & es1+es2 & .850 \\
    es2\texttt{(BE5)}  & es2\texttt{(BE5)} & .783 & es1+es2 & .820 \\
    de1\texttt{(VA)} & de1\texttt{(BE5)} & .825 & de1+de2 & .835 \\
    de2\texttt{(BE5)} & de2\texttt{(BE5)} & .780 & de1+de2 & .792 \\
    pl1\texttt{(VA)} & pl1\texttt{(BE5)} & .794 & pl1+pl2 & .841 \\
    pl2\texttt{(BE5)} & pl2\texttt{(BE5)} & .784 & pl1+pl2 & .835 \\
    tr1\texttt{(VA)} & tr1\texttt{(BE5)} & .600 & tr1+tr2 & .611 \\
    tr2\texttt{(BE5)} & tr2\texttt{(BE5)} & .613 & tr1+tr2 & .628 \\
    \hline
    Mean & \multicolumn{2}{r|}{.769} & \multicolumn{2}{r|}{.790} \\
    Disk Use & \multicolumn{2}{r|}{4.33 MB} & \multicolumn{2}{r|}{2.52 MB} \\
    \hline
\end{tabular}
\caption{Validation word-level results in the supervised setting.}
\label{tab:dev-result-word-supervised}
\end{table}

\begin{table}[t]
\centering
\footnotesize
\begin{tabular}{|l|l|r|l|r|}
    \hline
       & \multicolumn{2}{c|}{\textbf{Base Model (BERT)}} & \multicolumn{2}{c|}{\textbf{Base Model + PPH}} \\
    Test Data & Train Data & $r$ & Train Data & $r$ \\
    \hline
    EmoB & EmoB & .610 & EmoB+AffT & .600 \\
    AffT & AffT & .783 & EmoB+AffT & .790 \\
    CVAT & CVAT & .748 & CVAT & .749 \\
    \hline
    Mean & \multicolumn{2}{r|}{ .714} & \multicolumn{2}{r|}{ .713} \\
    {Disk Use} & \multicolumn{2}{r|}{1.25 GB} & \multicolumn{2}{r|}{0.81 GB} \\ 
    \hline
\end{tabular}
\caption{Validation text-level results in the supervised setting.}
\label{tab:dev-result-text-supervised}
\end{table}

\begin{table}[t]
\centering
\footnotesize
\begin{tabular}{|l|l|r|l|r|}
    \hline
        & \multicolumn{2}{c|}{\textbf{Base Model (FFN)}} & \multicolumn{2}{c|}{\textbf{Base Model + PPH}} \\
    Test Data & Train Data & $r$ & Train Data & $r$ \\
    \hline
    en1\texttt{(VAD)} & en2\texttt{(BE5)} & .762 & en2 & .778 \\ 
    en2\texttt{(BE5)} & en1\texttt{(VAD)} & .814 & en1 & .815 \\
    es1\texttt{(VA)} & es2\texttt{(BE5)} & .759 & es2 & .758 \\
    es2\texttt{(BE5)} & es1\texttt{(VA)} & .767 & es1 & .779 \\
    de1\texttt{(VA)} & de2\texttt{(BE5)} & .692 & de2 & .672 \\
    de2\texttt{(BE5)} & de1\texttt{(VA)} & .696 & de1 & .696 \\
    pl1\texttt{(VA)} & pl2\texttt{(BE5)} & .806 & pl2 & .829 \\
    pl2\texttt{(BE5)} & pl1\texttt{(VA)} & .776 & pl1 & .796 \\
    tr1\texttt{(VA)} & tr2\texttt{(BE5)} & .556 & tr2 & .571 \\
    tr2\texttt{(BE5)} & tr1\texttt{(VA)} & .556 & tr1 & .565 \\
    \hline
    Mean & \multicolumn{2}{r|}{.719} & \multicolumn{2}{r|}{.726} \\
    Method  & \multicolumn{2}{r|}{ext. post-processor} & \multicolumn{2}{r|}{built-in} \\
    \hline
\end{tabular}
\caption{Validation word-level results in the zero-shot setting.}
\label{tab:dev-result-word-zeroshot}
\end{table}

\begin{table}[t]
\centering
\footnotesize
\begin{tabular}{|l|l|r|l|r|}
    \hline
      & \multicolumn{2}{c|}{\textbf{Base Model (BERT)}} & \multicolumn{2}{c|}{\textbf{Base Model + PPH}} \\
    Test Data & Train Data & $r$ & Train Data & $r$ \\
    \hline
    EmoB & AffT & .353 & AffT & .368 \\
    AffT & EmoB & .636 & EmoB & .664 \\
    \hline
    Mean & \multicolumn{2}{r|}{ .495} & \multicolumn{2}{r|}{ .516} \\
     Method & \multicolumn{2}{r|}{ext. post-processor} & \multicolumn{2}{r|}{built-in} \\
    \hline
\end{tabular}
\caption{Validation text-level results in the zero-shot setting.}
\label{tab:dev-result-text-zeroshot}
\end{table}

Tables \ref{tab:dev-result-word-supervised} -- \ref{tab:dev-result-text-zeroshot} show the dev set results corresponding to the test set results in Tables \ref{tab:results-supervised-word} -- \ref{tab:results-zeroshot-text}, respectively. As can be seen, the former are consistent with the latter, yet overall slightly higher, as is usually the case.

\subsection{Evaluation Metric}

Prediction quality is evaluated using Pearson correlation defined as 

\begin{equation*}
r_{x,y} \coloneqq \frac{\sum_{i=1}^n (x_i-\bar{x})(y_i-\bar{y})}{\sqrt{\sum_{i=1}^n(x_i-\bar{x})^2} \; \sqrt{\sum_{i=1}^n(y_i-\bar{y})^2}}
\end{equation*}

where $x = x_1, x_2, \dots, x_n$, $y = y_1, y_2, \dots, y_n$ are real-valued number sequences and $\bar{x}$, $\bar{y}$ are their respective means. We rely on the implementation provided in the \textsc{SciPy} package.\footnote{\url{https://docs.scipy.org/doc/scipy/reference/generated/scipy.stats.pearsonr.html}}

\subsection{Model and Hyperparameter Selection}

As described in \sec\ref{sec:setup}, we mostly relied on hyperparameter choices by the authors of our base models. Hence, we performed only a relatively small amount of tuning throughout this work.

For the word base model and the label encoder, no further hyperparameter selection was required. For the text base model (BERT), we verified via a first round of development experiments that default settings yield satisfying prediction quality on our datasets. The learning rate of the \textsc{AdamW} optimizer was set to $10^{-5}$ based on established recommendations. Besides the number of training epochs (see below), the only dataset-specific hyperparameter choice had to be made for the batch size which we set according to constraints in GPU memory. (The samples in the CVAT dataset are significantly longer than in \textsc{AffT} so that fewer samples of the former can be placed in one batch.) We used the pre-trained weights "bert-base-uncased" and "bert-base-chinese" from \citet{Wolf20arxiv} for the English and Mandarin datasets, respectively. The dimensionality of the emotion space $\mathbb{R}^d$ was initially set to 100 and remained unchanged after verifying that the Multi-Way Mapping Model indeed showed good label mapping performance.

For each (word or text) dataset, we trained the models well beyond convergence, recording their dev set performance after each epoch  (number of epochs differs between datasets). We then chose the best-performing checkpoint (according to Pearson correlation) for the final test set evaluation.

Hyperparameter choices were identical between base models with and without PPH. We emphasize that for each base model, hyperparameters were set (by us or by the respective authors) with respect to base model \textit{without} PPH, thus forming a challenging testbed for our approach. We see an extensive hyperparameter search as a fruitful venue for future work.

\subsection{Data Access}

Below, we list URLs for all datasets used in our experiments.

\begin{description}
\footnotesize
\item[en1]  \url{https://osf.io/2k97q/download} (ratings must be extracted from PDF)
\item[en2] \url{https://static-content.springer.com/esm/art%3A10.3758%2FBF03192999/MediaObjects/Stevenson-BRM-2007.zip}
\item[es1] \url{https://static-content.springer.com/esm/art%3A10.3758%2Fs13428-015-0700-2/MediaObjects/13428_2015_700_MOESM1_ESM.csv}
\item[es2] \url{https://static-content.springer.com/esm/art%3A10.3758%2Fs13428-017-0962-y/MediaObjects/13428_2017_962_MOESM1_ESM.csv}
\item[de1] \url{https://www.ewi-psy.fu-berlin.de/einrichtungen/arbeitsbereiche/allgpsy/Download/BAWL/index.html}
\item[de2] \url{https://static-content.springer.com/esm/art%3A10.3758%2Fs13428-011-0059-y/MediaObjects/13428_2011_59_MOESM1_ESM.xls}
\item[pl1] \url{https://static-content.springer.com/esm/art%3A10.3758%2Fs13428-014-0552-1/MediaObjects/13428_2014_552_MOESM1_ESM.xlsx}
\item[pl2] \url{https://doi.org/10.1371/journal.pone.0132305.s004}
\item[tr1] \url{https://osf.io/rxtdm}
\item[tr2] \url{https://osf.io/rxtdm}
\item[\textsc{AffT}] \url{http://web.eecs.umich.edu/~mihalcea/affectivetext/}
\item[\textsc{EmoB}] \url{https://github.com/JULIELab/EmoBank}
\item[\textsc{CVAT}] \url{http://nlp.innobic.yzu.edu.tw/resources/cvat.html}
\end{description}

\subsection{Details of Train-Dev-Test Splits}

\textsc{EmoB} comes with a stratified split with ratios of about 8-1-1 (exactly 8062 train, 1000 dev, 1000 test samples). Since the samples of \textsc{AffT} are mostly also included in \textsc{EmoB}, we decided to use the data split of the latter for the former, too.  Samples of \textsc{AffT} that were not included in \textsc{EmoB} (about 5\% of the data) were removed before the experiments. CVAT features a 5-fold data split but without assigning the resulting parts to train, dev, or test utilization. We used the first three for training, the fourth for development/validation, and the fifth for testing.
 
 The word datasets in \tab\ref{tab:word-datasets} do not come with a fixed data split. Instead, we defined splits ourselves with ratios ranging between 3-1-1 to 8-1-1, depending on the number of samples. Instances were randomly assigned to train, dev, and test split using fixed random seeds. The resulting partitions were stored as JSON files and placed under version control.

\end{document}